\documentclass[10pt,twocolumn,letterpaper]{article}

\usepackage{cvpr}              % To produce the CAMERA-READY version
\definecolor{cvprblue}{rgb}{0.21,0.49,0.74}

\usepackage[pagebackref,breaklinks,colorlinks,allcolors=cvprblue]{hyperref}

\usepackage{graphicx}
\usepackage{amsmath}
\usepackage{amssymb}
\usepackage{booktabs}
\usepackage{multirow}
\usepackage{enumitem}
\usepackage[title]{appendix}
\usepackage{makecell}
\usepackage{amsmath}
\usepackage{amssymb}
\usepackage{bm}

\newcommand{\ours}{CAPT}
\usepackage[table,xcdraw]{xcolor}
\usepackage[dvipsnames, svgnames, x11names]{xcolor}
\definecolor{lightgray}{gray}{0.85}  % define a light gray color
\usepackage[pagebackref,breaklinks,colorlinks]{hyperref}

\usepackage[capitalize]{cleveref}
\crefname{section}{Sec.}{Secs.}
\Crefname{section}{Section}{Sections}
\Crefname{table}{Table}{Tables}
\crefname{table}{Tab.}{Tabs.}

\def\paperID{} % *** Enter the Paper ID here
\def\confName{CVPR}
\def\confYear{2025}

\title{\ours: Confusion-Aware Prompt Tuning for Reducing Vision-Language Misalignment}

\author{
  Maoyuan Shao\textsuperscript{1*},~~ 
  Yutong Gao\textsuperscript{1*†},~~
  Xinyang Huang\textsuperscript{2},~~
  Chuang Zhu\textsuperscript{2},~~
  Lijuan Sun \textsuperscript{3†},~~
 Guoshun Nan\textsuperscript{4}\\
  \textsuperscript{1}School of Information Engineering, Minzu University of China \\
  \textsuperscript{2}School of Artificial Intelligence, Beijing University of Posts and Telecommunications\\
  \textsuperscript{3} National Library of China, Beijing, China\\
  \textsuperscript{4} School of Cyberspace Security, Beijing University of Posts and Telecommunications\\
  \texttt{\small{\{ytgao92,maoyuanshao\}@muc.edu.cn, \{chuangzhu,xinyanghuang,sunlijuan\}@bupt.edu.cn}}
}

\begin{document}

\maketitle
\renewcommand\thefootnote{} % 去掉编号，让我们手动控制符号
\footnotetext{* Contribute  equally to this work.~~~
 † Corresponding authors.}
% \footnote{† Corresponding authors.}
% \footnote{* These authors contributed equally to this work.}
%%%%%%%%% ABSTRACT
\begin{abstract}
Vision-language models like CLIP have achieved remarkable progress in cross-modal representation learning, yet suffer from systematic misalignment among visually and semantically similar categories.
We observe that such confusion patterns are not random but persistently occur between specific category pairs, revealing the model’s intrinsic bias and limited fine-grained discriminative ability.
To address this, we propose CAPT, a \textbf{C}onfusion-\textbf{A}ware \textbf{P}rompt \textbf{T}uning framework that enables models to learn from their own misalignment.
Specifically, we construct a Confusion Bank to model stable confusion relationships across categories and their misaligned samples explicitly.
On this basis, we introduce a Semantic Confusion Miner (SEM) to capture global inter-class confusion through semantic difference and commonality prompts, and a Sample Confusion Miner (SAM) to retrieve representative misclassified instances from the bank and capture sample-level cues through a Diff-Manner Adapter that integrates global and local contexts.
To further unify confusion information across different granularities, a Multi-Granularity Discrepancy Expert (MGDE) module is designed to jointly leverage semantic and sample level experts for more robust confusion-aware reasoning.
Extensive experiments on 11 benchmark datasets demonstrate that our method significantly reduces confusion-induced errors while enhancing the discriminability and generalization of both base and novel classes, successfully resolving 50.72\% of confusable sample pairs. 
Code will be released at \href{https://github.com/greatest-gourmet/CAPT}{{https://github.com/greatest -gourmet/CAPT}}.
\end{abstract}
\vspace{-1.50em}

%%%%%%%%% BODY TEXT
\section{Introduction}
\label{intro}
 \begin{figure} 
\centering  
\includegraphics[width=1\linewidth]{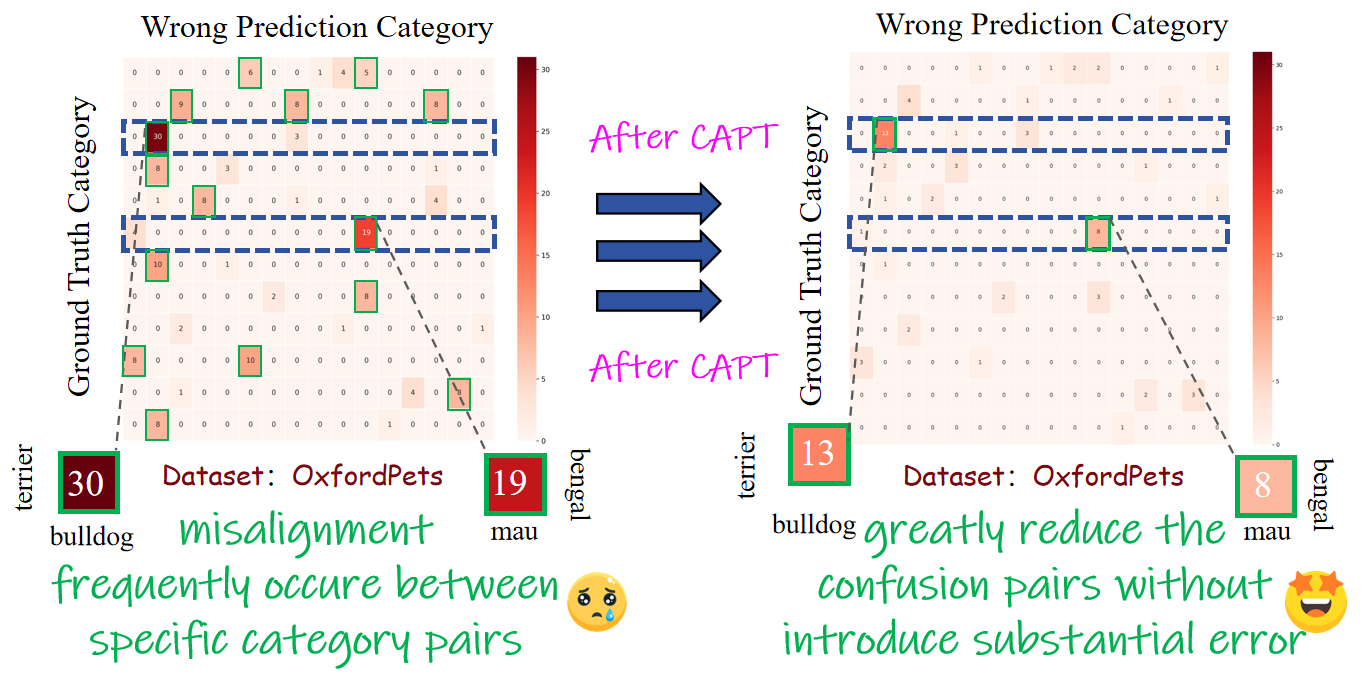} 
\caption{Heatmap of model misclassifications shows certain categories are consistently and frequently mispredicted as specific others, \textit{e.g.}, in OxfordPets, terrier is misclassified as bulldog 30 times while rarely being mistaken for other classes. CAPT significantly reduces such confusion rates, thereby improving overall accuracy.}
\label{fig:com}  
\vspace{-1em}
\end{figure}

%  \begin{figure} 
% \centering  
% \includegraphics[width=1.0 \linewidth]{./cot.png} 
% \vspace{-1.7em}  
% \caption {Construction of Semantic-level difference and commonality prompts. Through CoT leveraging model misclassification information, we progressively guide Large-Language-Module to generate more fine-grained prompts.}  
% \label{fig:cot}  
% \vspace{-1em}
% \end{figure}
Vision-language models like CLIP~\cite{RadfordKHRGASAM21} excel at learning unified visual-textual representations, providing a strong foundation for cross-modal alignment.
Building upon this capability, prompt tuning enables efficient adaptation to diverse downstream tasks by introducing learnable prompts that steer cross-modal interactions toward task-relevant semantics~\cite{ZhouYLL22,ZhouYL022}.
This approach produces more discriminative joint embeddings and exhibits strong generalization across tasks such as open-domain recognition~\cite{hu2023open}, human-computer interaction~\cite{mao2023clip4hoi}, and multimedia retrieval~\cite{hu2023adaclip}.

To further enhance cross-modal alignment, recent studies~\cite{tian2024learning,alonso2025vision} have primarily focused on optimizing the interaction mechanisms between visual and textual modalities.
Some methods~\cite{guo2025parameter,ZhouYLL22,chen2024revisiting} employ learnable prompt tokens to capture global semantic relationships, while others~\cite{kim2024efficient,xie2025textrefiner} introduce additional adapters or attention modules to strengthen modality fusion.
These approaches have brought substantial improvements to the overall quality of cross-modal embeddings, leading to more robust and semantically consistent representations across modalities.

However, as shown in Figure~\ref{fig:com}, despite their strong overall performance, these models exhibit systematic misalignment on challenging samples. Certain categories are consistently misaligned into specific target classes, with confusion probabilities much higher than for other classes; for instance, on the OxfordPets dataset, terrier instances were misaligned with bulldog 30 times, while rarely being confused with any other classes.
This persistent confusion pattern indicates that the model struggles to distinguish between visually and semantically similar classes, reflecting an underlying vision–language misalignment where the textual and visual embeddings fail to capture subtle intra-class differences.
Such misalignment constrains the model’s ability to perform fine-grained semantic discrimination, ultimately limiting its robustness and generalization.

To address the aforementioned issue, we revisit prompt tuning from the perspective of confusable category differentiation and introduce a confusion-aware prompt tuning framework (CAPT). 
By explicitly modeling the relationships among easily confusable categories and their misaligned samples, CAPT encourages the model to identify and rectify vision–language discrepancies that lead to systematic confusions and repeated misalignments, enhancing fine-grained discrimination on challenging instances.

Specifically, as model misalignment often arises from ambiguous semantic boundaries and local similarities among sample representations~\cite{wang2020learning,cheng2025confusion}, we explicitly mine confusions at two complementary levels: semantic and sample.
 Semantic Confusion Miner (SEM) focuses on capturing global inter-class confusion patterns. 
By leveraging cross-modal matching confidences and semantic confusion statistics, we simulate potential confusion behaviors and identify stable confusion modes between semantic confusion pairs, which are then distilled into semantic difference and commonality prompts, explicitly guiding the model to disentangle semantically confusing categories.
Meanwhile, Sample Confusion Miner (SAM) further captures fine-grained discrepancies at the instance level.
Given each sample and its semantic confusion pairs, we retrieve confused samples from the confusion bank and select the most representative confusion ones based on feature similarity.
Finally, through the Diff-Manner Adapter in Sample Confusion Miner, which autonomously fuses global and local cues, we extract the final sample confusion feature that captures instance-level confusion dynamics.

To further integrate confusion information across multiple granularities, we propose a  Multi-Granularity Discrepancy Expert (MGDE) module comprising semantic and sample level experts.
Each expert focuses on distinct aspects of confusion representation and feature alignment, collaboratively modeling from high-level semantic concepts to fine-grained instances to better reduce misalignment and enhance discrimination among confusing categories.
In addition, we cluster semantic prompt tokens to refine the embedding structure, yielding compact yet expressive prompts while mitigating the influence of low-discriminative tokens on confusing category prediction.
% To further efficiently integrate confusion information across different granularities, we first introduce a Difference-Aware Modulator, which leverages semantic-level prompts to guide the model in adaptively switching between semantic commonality understanding and sample-level feature perception, directing attention to critical differences within the confusion-contrast set.
% Next, we design a Multi-Granularity Difference Experts module, composed of semantic-level and sample-level experts, focusing on confusion representations and feature guidance at their respective levels.
% This enables collaborative modeling from semantic concepts to individual samples, enhancing the model’s sensitivity to subtle inter-class differences.
% Additionally, we cluster text tokens to refine the structure of text embeddings, thereby compressing the prompt length and mitigating the interference of low-discriminative tokens in confusable category prediction.

Experiments on 11 datasets demonstrate that CAPT substantially reduces vision–language misalignment among confusable samples without introducing additional issues.
Compared with previous approaches,  CAPT consistently achieves higher performance on both base and novel classes, attaining 87.41\% and 80.90\% accuracy, and resolving 50.72\% of confusable sample pairs, suggesting that  CAPT can effectively guide the model to learn from misalignment and better distinguish confusing categories.

Our main contributions can be summarized as follows:
\begin{itemize}
 \item We identify a fixed confusion pattern in model misalignment and propose CAPT, which explicitly models the relations between confusable classes and trained samples, enabling learning from its own misalignment.
 \item 
 We characterize confusion relationships at both the semantic (SEM) and sample (SAM) levels, and fuse multi-level fine-grained confusion cues through the proposed MGDE, enabling the model to more effectively capture and learn diverse forms of confusion.
  \item We significantly reduce the misalignment rate of confusable samples and enhance the cross-modal alignment and generalization of novel categories, achieving the best accuracy of 83.90\% on Harmonic Mean (HM).
\end{itemize}

\section{Related Works}
\label{sec:related}      %短一点啊
\noindent\textbf{Vision-Language Models.}
Large-scale Vision-Language Models (VLMs) jointly encode visual and textual modalities using massive image–text datasets, achieving strong generalization and discriminative capabilities~\cite{RadfordKHRGASAM21,abs-2204-14198}. Representative models such as CLIP~\cite{RadfordKHRGASAM21}, MLM~\cite{KimSK21,LuBPL19}, MAE~\cite{HeCXLDG22} and LFA~\cite{BlackBox} typically adopt contrastive or masked-prediction objectives to learn robust multimodal representations. For example, CLIP aligns large-scale image–text pairs through contrastive learning, while MLM and MAE~\cite{HeCXLDG22} enhance representation learning via random masking. 
Despite their success, adapting pre-trained VLMs to downstream tasks, especially under few-shot scenarios, remains challenging, motivating research on improved encoders~\cite{VaswaniSPUJGKP17,zhai2022scaling,li2023blip}, deeper modality integration~\cite{li2022blip,singh2022flava}, and larger-scale training~\cite{RadfordKHRGASAM21,JiaYXCPPLSLD21,schuhmann2021laion,schuhmann2022laion}.
In this work, we propose  CAPT for prompt tuning CLIP by learning from its  
misalignment errors through confusion pairs.

\noindent\textbf{Prompt Tuning for VLMs.}
Prompt learning adapts pre-trained models to downstream few-shot tasks by reformulating them with prompts, bridging domain gaps and leveraging prior knowledge. 
Early approaches~\cite{ZhouYLL22,abs-2205-14865,abs-2210-07225} relied on manually crafted templates, while later works such as MaPLe~\cite{MAPLE23} and PromptSRC~\cite{DBLP:journals/corr/abs-2307-06948} jointly aligned visual and textual prompts. Adapter-based methods~\cite{DBLP:conf/eccv/ZhangZFGLDQL22,abs-2110-04544,kim2024efficient,xie2025textrefiner} further introduced context-aware adaptation through lightweight transformer adapters.
In addition, Skip Tuning~\cite{wu2025skip} improves training efficiency by reducing both the length and width of feature gradients, while Spotlighter~\cite{gao2025spotlighter} accelerates inference by pruning redundant tokens.
Despite their success, these methods overlook the “fixed confusion patterns” shared across models, leading to repetitive misalignment among similar samples. In this work, we exploit confusion information at both the semantic and sample levels to help the model better distinguish confusing categories and enhance fine-grained discriminative ability.
% Adapting a pre-trained model to a downstream task usually requires fine-tuning all parameters, which can lead to large model sizes and overfitting.
% To address this issue, prompt tuning~\cite{ZhouYLL22,abs-2205-14865,abs-2210-07225} introduces task-specific tags to better leverage task-related knowledge.
% CoOp~\cite{ZhouYLL22} enhances CLIP by optimizing continuous prompt vectors only within the text encoder.
% CoCoOp~\cite{ZhouYL022} and VPT~\cite{abs-2210-07225} extend this by conditioning prompts on specific image instances, while KgCoOp~\cite{YaoZX23}, ProGrad~\cite{abs-2205-14865}, and ProReg~\cite{PromptReg23} introduce constraints to maintain necessary general knowledge.
% Recent methods~\cite{ECO23,proda,DAPT23,KAPT23,0002YSLR023} have further improved visual-language alignment by maximizing inter-class dispersion.
% For the multi-encoder setting, MaPLe~\cite{MAPLE23} and PromptSRC~\cite{DBLP:journals/corr/abs-2307-06948} focus on joint visual-textual prompt alignment, while MVLPT~\cite{DBLP:journals/corr/abs-2211-11720} incorporates cross-task knowledge. Adapter-based methods~\cite{DBLP:conf/eccv/ZhangZFGLDQL22,abs-2110-04544,kim2024efficient} use context-aware prompts for dense prediction tasks by introducing adapters after each self-attention layer.
% In this work, we focus on optimizing multi-modal prompts and propose MoDe by introducing priors to improve existing works lacking multi-modal discriminative knowledge.

%只是一个简述，思路很乱，还没有整理,只是告诉你这里边有什么，突然发现这边公式不统一
\begin{figure*}
 \centering
  \includegraphics[width=1\linewidth]{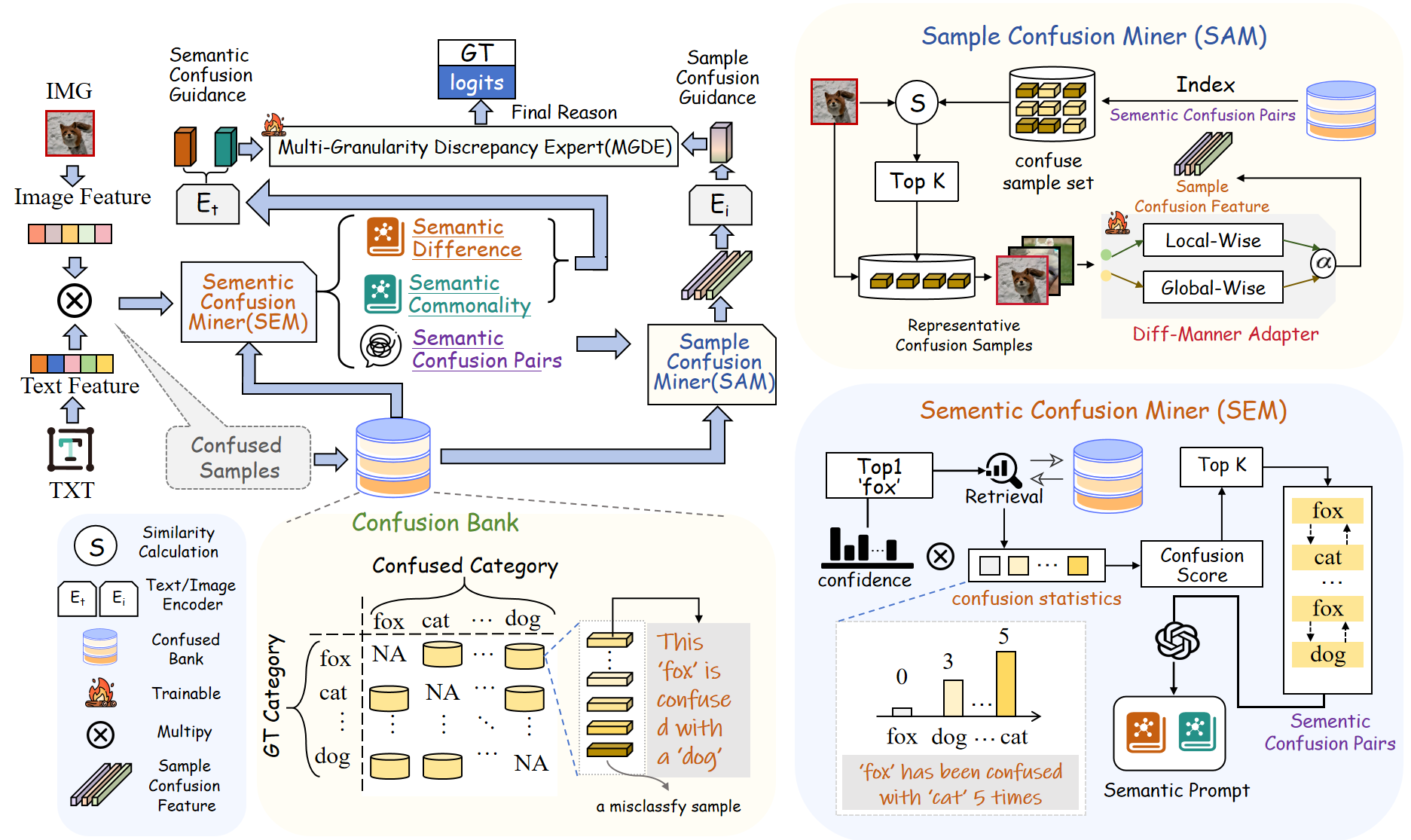}
  \caption{Overview of CAPT. By matching feature representations, we first employ a Semantic Confusion Miner (SEM) that, together with statistics from the Confusion Bank, identifies Semantic Confusion Pairs and generates both commonality and difference prompts. Subsequently, the Sample Confusion Miner (SAM) locates the most representative confusing samples based on these pairs and extracts their Sample Confusion Feature via the Diff-Manner Adapter. Finally, the Multi-Granularity Discrepancy Expert (MGDE) module integrates semantic and sample level confusion information for unified representation refinement. Framework of MGDE is shown in Figure~\ref{fig:mgde}.}
  \label{fig}
\end{figure*}
\section{Methodology}
We introduce CAPT, a framework that encourages the model to discover and correct its own misalignment errors through confusion-aware difference mining in prompt tuning.
Figure \ref{fig} shows our overall learning process.
We first review CLIP and loss for Confusion-Aware Prompt Tuning in Section \ref{sec1} and then introduce the whole work in detail.
\subsection{Preliminary}
\label{sec1}
\noindent\textbf{Vision-Language Models.}
Our model is built upon CLIP, which employs contrastive learning to align representations across visual and textual space, thereby enabling strong generalization in both few-shot and zero-shot scenarios.
Given a labeled image classification dataset $\mathcal{D}={(x,c)}$ with $N$ classes $\mathcal{C}={C}_{i=1}^N$, CLIP predicts labels by computing the cosine similarity between image features $f_I$ and the text features $w$ of each class.
Concretely, for each image $x$, the image encoder $E_I(x)$ extracts a feature vector $f_I = E_I(x)$, while class-specific hand-crafted prompts $t$ (e.g., “a photo of a [cls]”) are encoded by the text encoder $E_T(t)$ to obtain text features $w = E_T(t)$.
Finally, the probability of classifying an image $x$ into class $c$ is given by:
\begin{equation}
{
P(c \mid x) = \frac{\exp\left(\cos\!\left( f_I,  w_c\right)/\tau\right)}
{\sum\limits_{c' \in \mathcal{C}} \exp\left(\cos\!\left( f_I,  w_c'\right)/\tau\right)},}
\label{equ1}
\end{equation}
where $\tau$ is the temperature parameter and 
$\cos(\cdot,\cdot)$ denotes the cosine similarity.

\noindent\textbf{Loss for Confusion-Aware Prompt Tuning.}
During prompt tuning, the training objective focuses solely on aligning global image–text features, optimizing contrastive learning to enhance the similarity of positive pairs as:
\begin{equation}
    \mathcal{L}_{ori} = -\sum_{j}  {y}_j \log \mathrm{p}(c_j |  {x}_i),
\end{equation}
where $y$ is the one-hot label.
As confusion samples are introduced at multiple levels, a simple loss fails to capture the fine-grained inter-class confusion.
Inspired by DPC~\cite{li2025dpc} and TripletCLIP~\cite{patel2024tripletclip}, an InfoNCE~\cite{parulekar2023infonce} loss can better exploit misaligned information, thereby enhancing discriminative confusion learning among selected samples.
For the final features  $f_I$ of confused samples and $f_T$ of prompts with $L$ pair length, the final loss is optimized as follows:
\begin{equation}
    \mathcal{L}_{confuse} = -\frac{1}{L} \sum_{i=1}^{L} \left( \log p_\theta \left( y \mid f_I \right) + \log p_\theta \left( y \mid f_T \right) \right),
    \label{loss1}
\end{equation}
where $p_\theta\left(y \mid  {f}_I\right)$ measures the compatibility between the training image and the text feature, the symmetric term captures the reverse compatibility.
\begin{equation}
\small
p_\theta(y \mid f_I) = \frac{\exp\left(\text{sim}(f_T, f_I) / \tau\right)}{\sum_{j=1}^{L} \exp\left(\text{sim}(f_T, f_I^{(j)}) / \tau\right)}.
\label{yfi}
\end{equation}
\begin{equation}
\small
p_\theta(y \mid f_T) = \frac{\exp\left(\text{sim}(f_I, f_T) / \tau\right)}{\sum_{j=1}^{L} \exp\left(\text{sim}(f_I, f_T^{(j)}) / \tau\right)}.
\label{rev}
\end{equation}

\subsection{Confusion-Aware Prompt Tuning}
% To effectively extract confusion-aware discrepancies from the model’s misclassifications, we employ both a semantic confusion miner and a sample-level confusion miner by capturing the interactions between the current sample and confusable categories. We further introduce a Multi-Granularity Difference Expert module to fuse semantic- and sample-level cues, enabling reasoning over both global semantics and local instance patterns. This enhances the model’s ability to distinguish highly similar categories and resolve complex confusion cases.  
% We elaborate on their functionalities in the following sections.
Misaligned samples encapsulate informative confusion patterns, reflecting the model’s limited fine-grained discrimination while providing valuable signals for refinement through prompt tuning. 
Analyses across multiple datasets reveal consistent, class-specific confusion patterns that expose systematic misalignment and highlight inherent biases in model recognition.
To systematically organize the confusion pattern, we first build a Confusion Bank that records each sample under the category it is misalignment into, forming an index of inter-class confusion relationships.
Within the Bank, the Semantic Confusion Miner (SEM)  facilitates a systematic understanding of the model’s semantic level confusion patterns, and the Sample Confusion Miner (SAM) retrieves truly confusing samples, steering the model to learn the most representative confusion feature.
Finally, we integrate these two types of information via the Multi-Granularity Discrepancy Expert (MGDE). 
A detailed description of each module is presented below.

\noindent\textbf{Semantic Confusion Miner.}
To index semantic confusion pairs in the Confusion Bank, a straightforward strategy is to use the manually annotated ground-truth category as GT.
However, this may not accurately reflect the model’s intrinsic confusion tendencies. Instead, for each image–text pair, we first compute its confidence distribution $C$ over all categories and define the pseudo-GT as:
\begin{equation}
\textit{pseudo-GT} = \arg\max_j C_j,
\end{equation}  
where the pseudo-GT corresponds to the category with the highest confidence predicted by the pretrained model, serving as a surrogate for the ground truth. This design simulates the model’s latent confusion behavior and aligns subsequent confusion modeling with its inherent semantic bias, thereby better capturing semantically confusable relationships.
Meanwhile, directly selecting the remaining top-k categories based on confidence to construct semantic confusion pairs tends to overemphasize the current sample while neglecting the overall semantic distribution.
To this end, we retrieve the confusion statistics of the pseudo-GT from the Confusion Bank, recording the number $n_i$ of misclassified samples for the \textit{i-th} category.
These statistics are then normalized and integrated with the current sample’s confidence to compute the final confusion score $S$, as follows:
\begin{equation}
    S_i = (1 + \frac{n_i}{\sum_{i=1}^{c} n_i})C_i,
    \label{score}
\end{equation}
where the “+1” term is added to ensure numerical stability when a category is absent from the pseudo-GT's Confusion Bank.
To further enhance the modeling of semantic confusion relationships, inspired by CoT~\cite{wei2022chain,li2024atprompt}, we leverage LLM to generate fine-grained semantic prompts for the Semantic Confusion Pairs, capturing both their commonality and differences, with the process shown in the Appendix.
These enriched semantic prompts facilitate the model's learning of differences and commonalities.

\noindent\textbf{Sample Confusion Miner.}
To further exploit sample-level confusion cues, we retrieve the most representative misclassified samples for each instance to characterize its local confusion structure within the feature space.
Relying solely on semantic cues overlooks subtle intra-class confusion, and incorporating sample level confusion information provides complementary fine-grained signals that help the model better separate visually similar categories.
% In contrast to prior random selection strategies, our approach adaptively mines confusion instances that are highly correlated with the current sample, enabling more precise discrimination among visually similar categories and more effective correction of historical misclassifications.
Specifically, guided by the previously derived semantic confusion pairs, we query the Confusion Bank to obtain a confused sample set consisting of misclassified samples from each confusing category.
To mitigate noise introduced by redundant samples, we retain only the most representative confusing sample for each confusion category, ensuring semantic representativeness while enabling more efficient and robust modeling.
Therefore, we denote the confused sample set as ${U}\in\mathbb{R}^{c \times l}$, where $c$ represents the number of confusion pairs and $l$ denotes the number of misclassified samples within each confused category.
We then compute the feature similarity between $U$ and the current instance $I$ to identify the most representative confusing samples $I_c^{*}$ for each confusion pairs in:
\begin{equation}
I_c^{*} = \arg\max_{j \in c,i \in l} \cos(E_I(I), E_I(U_j^{i})),
\label{sim}
\end{equation}
where $E_I$ is the image encoder.
Unlike random selection strategies~\cite{li2025dpc}, deliberately choosing misclassified samples enables targeted mining of confusion instances closely related to the current sample, thereby improving the model’s ability to distinguish similar categories and correct prior misalignments.
Then, to better capture the confusion patterns between $I_c^{*}$ and $I$, we introduce a \textbf{Diff-Manner Adapter}, jointly integrating local and global difference cues.
Prior works
~\cite{dosovitskiy2020image,raghu2021vision,guo2022cmt} 
show that ViTs capture global context to identify commonalities among confusing samples, while convolutions emphasize local details to distinguish sample confusion differences.
Accordingly, we integrate convolutional modules into the standard ViT architecture and employ a dynamic weight $\alpha$ to allow the model to adaptively determine which cues to emphasize.
Specifically, let \(Y \in \mathbb{R}^{(N+1)\times D}\) denote a block of the image encoder, where the input is given by \([c; X] \). Here, \(c\) denotes the additional $cls$ token, \(D\) represents the hidden embedding dimension of each token and \(X = [x_j]_{j=1}^N\) consists of the embeddings of \(N\) image patches.
We first apply an attention mechanism between the $[CLS]$ token and the image patches, emphasizing the perception of global features:
\begin{equation}
\small
   \hat{[X]} = \text{MultiHead} (\text{LN} ([CLS]), \text{LN} ({  [X]}),\text{LN} ({[X]}))  .
   \label{eq_x}
\end{equation}
 Subsequently, the sequential patch tokens are reshaped into a tensor $R^{R\times C \times D}$, and a 2D depthwise convolutional layer is applied  to the extracted image patches along the spatial dimensions, aiming to capture local details and further combine with global information through a dynamic weight :
\begin{equation}
\small
 {[X]} \leftarrow {[X]} + \alpha \cdot DWConv2D(\hat{[X]}).
 \label{eq3}
\end{equation}
The final sample confusion  feature combines the instance’s local differences, global context, and confusion features to comprehensively capture instance-level confusion patterns for effective confusion-aware learning.

\begin{figure} 
\centering  
\includegraphics[width=0.8\linewidth]{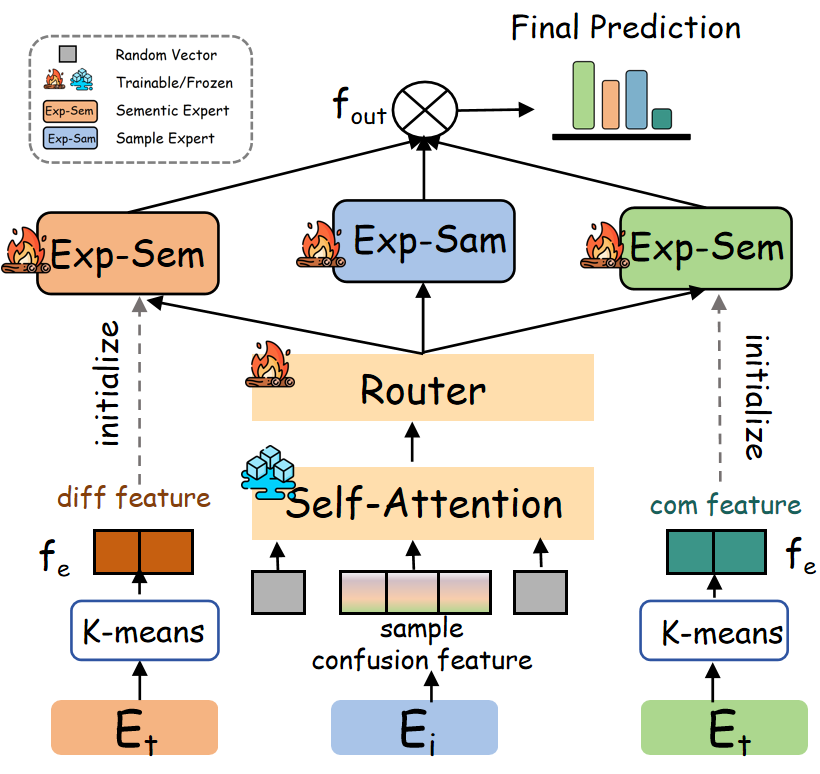} 
\caption{Overview of Multi-Granularity Discrepancy Expert.
MGDE integrates semantic and sample level experts to fuse dual confusion cues, reinforced by random vectors and specific initialization to enhance confusion learning in prompt tuning.}
\label{fig:mgde}  
\vspace{-1em}
\end{figure}

\noindent\textbf{Multi-Granularity Discrepancy Expert.}
To effectively integrate semantic and sample level information, we propose the Multi-Granularity Difference Expert (MGDE) model, 
which enhances the model’s ability to capture confusion patterns across multiple granularities and discriminate visually similar categories, shown in Figure~\ref{fig:mgde}. 
Confused knowledge at different levels exhibits distinct representations and distributions, making conventional fusion methods prone to information loss or interference.
To address this, we design two dedicated experts to maintain learnable features at the semantic and sample levels, respectively, enabling collaborative modeling of confusion information across granularities. This approach effectively enhances the model’s ability to differentiate highly similar categories and improves its perception of complex confusion patterns.
Following a standard Mixture-of-Experts architecture~\cite{jiang2024mixtral}, the overall formulation is calculated as:
\begin{equation}
\small
\begin{split}
 {f}_{\text{out}} &= \sum_{i=1}^{K} \text{Softmax}(\text{TopK}( {f} \cdot  {W}_r)) \cdot Exp_i( {f}), 
\end{split}
\label{eq4}
\end{equation}
where $W_{r}$ denotes the randomly initialized routing network, $f$ represents the mined sample confusion feature, and $Exp$ corresponds to the sample-level and semantic-level experts.  
The sample-level expert is initialized from the FFN embedding of the original CLIP model, while the semantic-level expert is initialized from textual embeddings derived from semantic difference and commonality prompts.  
Notably, to mitigate the interference caused by low-discriminative prompt tokens in fine-grained classification and to reduce the dimensionality of the prompt embeddings, we perform clustering-based optimization on the prompts $t_{prompt}$ and project them into the expert space $f_e$ through a linear layer:
\begin{equation}
f_e =  {w}^T \cdot \textit{K-means}( {t}_{\text{prompt}}).
\label{eq5}
\end{equation}
This design yields more compact and discriminative semantic representations, enhancing the model’s capability to distinguish confusing categories and improving its generalization performance.
During training, random vectors are assigned to the activated experts, and masking is applied to their corresponding feature cues to guide the model in learning task-specific knowledge.
This process guides the model to focus selectively on expert features and become more sensitive to confusing information.
Notably, two semantic experts model common and distinct semantics dependently without parameter sharing; together with the sample expert, their outputs are adaptively fused by a lightweight router.

\begin{table*}[t] 
  
  \caption{Comparison with other methods on base-to-new generalization with 16-shot.} % 表格标题
  \label{tab:base-to-new-generalization}
  \centering
  \resizebox{\textwidth}{!}{ 
  \begin{tabular}{l|ccc|ccc|ccc|ccccccc} % l表示左对齐，后面的c表示居中对齐，根据实际列数确定
    \toprule
    \multirow{2}{*}{Method} & \multicolumn{3}{c}{Average} & \multicolumn{3}{c}{ImageNet} & \multicolumn{3}{c}{Caltech101} & \multicolumn{3}{c}{OxfordPets} \\
    \cmidrule (lr){2 - 4} \cmidrule (lr){5 - 7} \cmidrule (lr){8 - 10} \cmidrule (lr){11 - 13}
    & Base & Novel & HM & Base & Novel & HM & Base & Novel & HM & Base & Novel & HM \\
    \midrule
    CoOp \scriptsize{(IJCV 22)} & 82.69 & 63.22 & 71.66 & 76.47 & 68.78 & 71.92 & 98.00 & 89.81 & 93.73 & 93.67 & 95.29 & 94.47 \\
     MaPLe \scriptsize{(CVPR 23)} & 82.28 & 75.14 & 78.55 & 76.66 & 70.54 & 73.47 & 97.74 & 94.36 & 96.02 & 95.43 & 97.76 & 96.58 \\
    PromptKD \scriptsize{(CVPR 24)} & 86.96 & 80.73 & 83.73 &  \textbf{80.83} & 74.66 & 77.62 & 98.91 & \textbf{96.65} & 97.77 & 96.30 & \textbf{98.01} & \textbf{97.15}\\
     Spotlighter  \scriptsize{(EMNLP 25)} & 85.65 & 80.46 & 82.89 &77.62& 71.71  & 74.55  & 98.86  & 96.74  & 97.79 &\textbf{96.48} &97.75 &97.11 \\
         2SFS  \scriptsize{(CVPR 25)} &85.55 &75.48 &80.20 &77.71 &70.99 &74.20 &98.71 &94.43 &96.53 &95.32 &97.82 &96.55\\
         TAC  \scriptsize{(CVPR 25)}   &85.42 &77.60 &81.24 &78.57 &71.03 &74.61 &98.57 &95.27 &96.89 &95.93 &98.17 &97.04\\
          LwEIB  \scriptsize{(IJCV 25)}  &84.45 &78.21 &81.21 &76.64 &71.64 &74.06 &98.47 &95.47 &96.95 &95.70 &97.40 &96.54\\
          TAP    \scriptsize{(ICLR 25)} &84.75 &77.63 &81.04 &77.97 & 70.40 & 73.99 & 98.90 & 95.50 &97.17 &95.80 &97.73 &96.76\\
          \midrule
           \rowcolor{orange!10}
% \textit{CAPT} &\textbf{87.41}\textcolor{red}{\scriptsize{+0.45}}  &\textbf{80.90} \textcolor{red}{\scriptsize{+0.17}} &\textbf{83.90}\textcolor{red}{\scriptsize{+0.17}} & 80.51\textcolor{green}{\scriptsize{-0.32}} &\textbf{75.13}\textcolor{red}{\scriptsize{+0.47}} &\textbf{77.73}\textcolor{red}{\scriptsize{+0.11}} &\textbf{99.10} \textcolor{red}{\scriptsize{+0.19}}& 96.34\textcolor{green}{\scriptsize{-0.13}}&\textbf{97.70}\textcolor{red}{\scriptsize{+0.17}} & 96.44\textcolor{red}{\scriptsize{+0.13}} & 97.71\textcolor{green}{\scriptsize{-0.30}} & 97.07\textcolor{green}{\scriptsize{-0.08}}\\
\textbf{CAPT} &\textbf{87.41} &\textbf{80.90} &\textbf{83.90} & 80.51 &\textbf{75.13} &\textbf{77.73} &\textbf{99.10} & 96.34&\textbf{97.70} & 96.44 & 97.71& 97.07\\
   
    \midrule
    \multirow{2}{*}{Method} & \multicolumn{3}{c}{StanfordCars} & \multicolumn{3}{c}{Flowers102} & \multicolumn{3}{c}{Food101} & \multicolumn{3}{c}{FGVCAircraft} \\
    \cmidrule (lr){2 - 4} \cmidrule (lr){5 - 7} \cmidrule (lr){8 - 10} \cmidrule (lr){11 - 13}
    & Base & Novel & HM & Base & Novel & HM & Base & Novel & HM & Base & Novel & HM \\
    \midrule
    CoOp \scriptsize{(IJCV 22)} & 78.12 & 60.40 & 68.13 & 97.60 & 59.67 & 74.06 & 88.33 & 82.26 & 85.19 & 40.44 & 22.30 & 28.75 \\
        MaPLe  \scriptsize{(CVPR 23)}& 72.94 & 74.00 & 73.47 & 95.92 & 72.46 & 82.56 & 90.71 & 92.05 & 91.38 & 37.44 & 35.61 & 36.50 \\
    PromptKD  \scriptsize{(CVPR 24)}& 82.80 & 83.37 & 83.13 & \textbf{99.42} & 82.62 & 90.24 & 92.43 & 93.68 & 93.05 & 49.12 & \textbf{41.81} & \textbf{45.17} \\
     Spotlighter  \scriptsize{(EMNLP 25)} & 81.62  & 82.15 & 81.88  & 99.36 & \textbf{83.47}  & \textbf{90.72}  &91.86 & 92.93  &92.39 & 46.35  & 40.68  & 43.33 \\
     2SFS  \scriptsize{(CVPR 25)} &82.50 &74.80 &78.46 &98.29 &76.16 &85.83 &89.11 &91.34 &90.21 &47.48 &35.51 &40.63\\
     TAC  \scriptsize{(CVPR 25)}&81.63 &74.17 &77.72 &97.97 &76.87 &86.15 &90.87 &91.87 &91.37 &44.60 &37.70 &40.86\\
     LwEIB  \scriptsize{(IJCV 25)} &80.07 &74.01 &76.92 &97.53 &77.50 &86.37 &90.63 &91.73 &91.18 &45.11 &42.60 &43.82\\
      TAP    \scriptsize{(ICLR 25)}  &80.70 &74.27 &77.35 &97.90 &75.57 &85.30 &90.97 &91.83 &91.40 &44.40 &36.50 &40.06\\
      \midrule
      \rowcolor{orange!10}
\textbf{CAPT} &\textbf{83.43} &\textbf{84.26} &\textbf{83.84} & 98.78 & 82.91 & 90.15&\textbf{92.43} &\textbf{93.96} &\textbf{93.19} &\textbf{49.67} & 40.31& 44.50\\

     \midrule
    \multirow{2}{*}{Method} & \multicolumn{3}{c}{SUN397} & \multicolumn{3}{c}{DTD} & \multicolumn{3}{c}{EuroSAT} & \multicolumn{3}{c}{UCF101} \\
    \cmidrule (lr){2 - 4} \cmidrule (lr){5 - 7} \cmidrule (lr){8 - 10} \cmidrule (lr){11 - 13}
    & Base & Novel & HM & Base & Novel & HM & Base & Novel & HM & Base & Novel & HM \\
    \midrule
    CoOp \scriptsize{(IJCV 22)} & 80.60 & 65.89 & 72.51 & 79.44 & 41.18 & 54.24 & 92.19 & 54.74 & 68.69 & 84.69 & 56.05 & 67.46 \\
        MaPLe  \scriptsize{(CVPR 23)}& 80.82 & 78.70 & 79.75 & 80.36 & 59.18 & 68.16 &  94.07 & 73.23 & 82.35 & 83.00 & 78.66 & 80.77 \\
    PromptKD  \scriptsize{(CVPR 24)} & 83.69 & \textbf{81.54} & 82.60 & 85.84 & 71.37 & 77.94 & 97.54 & 82.08 & \textbf{89.14} & 89.71 & 82.27 & 86.10 \\
    Spotlighter  \scriptsize{(EMNLP 25)}& 83.15  & 81.06 & 82.09 & 83.94  & 71.92  & 77.47  &93.17 & \textbf{84.51} & 88.63  & 89.72  &82.16 & 85.77\\
    2SFS  \scriptsize{(CVPR 25)}  &82.59 &78.91 &80.70 &84.60 &65.01 &73.52 &96.91 &67.09 &79.29 &87.85 &78.19 &82.74\\
    TAC  \scriptsize{(CVPR 25)} &83.70 &80.03 &81.82 &83.37 &64.27 &72.58 &94.37 &82.60 &88.10 &88.07 &81.67 &84.75\\
    LwEIB  \scriptsize{(IJCV 25)}&81.10 &79.80 &80.44 &82.87 &67.83 &74.60 &95.00 &80.01 &86.86 &85.73 &82.37 &84.02\\
     TAP    \scriptsize{(ICLR 25)}  &82.87 &79.53 &81.17 &84.20 &68.00 &75.24 &90.70 &82.17 &86.22 &87.90&82.43 &85.08\\
\midrule
 \rowcolor{orange!10}
\textbf{CAPT} &\textbf{83.98} & 81.32 &\textbf{82.63} &\textbf{88.19} &\textbf{72.38} &\textbf{79.51} &\textbf{97.60} & 81.32&88.72 &\textbf{91.42} &\textbf{82.36} &\textbf{86.65}\\

    \bottomrule
  \end{tabular}
  }
\end{table*}

\begin{table*}[t]
\centering
\caption{Comparative results in the cross-dataset transfer setting. Bold accuracies are the highest. Optimization is conducted on ImageNet and evaluation is performed on other datasets.}
\resizebox{\linewidth}{!}{%
\begin{tabular}{lcccccccccccc}
\toprule
& \multicolumn{1}{c}{\textbf{Source}} & \multicolumn{10}{c}{\textbf{Target}} & \\
\cmidrule(lr){2-2} \cmidrule(lr){3-13}
& ImageNet & Caltech101 & OxfordPets & StanfordCars & Flowers102 & Food101 & FGVCAircraft & SUN397 & DTD & EuroSAT & UCF101 & \textit{Average} \\
\midrule
CoOp   & 71.51 & 93.70 & 89.14 & 64.51 & 68.71 & 85.30 & 18.47 & 64.15 & 41.92 & 46.39 & 66.55 & 63.88 \\
PromptKD &  -    &93.61 &91.59 &73.93 &75.33 &\textbf{88.84} &26.24 &68.57 &55.08 &63.74 &\textbf{76.39} &71.33\\
LwEIB   &71.31 &94.51 & 92.50 &66.58 &73.03 &86.37 &\textbf{27.70} &69.33 &50.63 &55.37 &70.03 &68.81\\
TAC   &\textbf{72.77} &94.53 &90.67 &65.30 &72.20 &85.83 &23.53 &67.63 &47.57 &48.07 &70.00 &66.53\\
\midrule
\rowcolor{orange!10}
 \textbf{CAPT} &72.69 &\textbf{94.62} &\textbf{92.53} &\textbf{74.37} &\textbf{75.41} &88.75 &27.23 &\textbf{69.43} &\textbf{57.21} &\textbf{64.74} &75.19 &\textbf{71.95}\\
\bottomrule
\end{tabular}
}
\label{tab:cross-dataset}
\end{table*}

\begin{table}[h]
\setlength{\tabcolsep}{4pt} 
\caption{Comparison with other methods on cross-domain generalization with 16-shot.}
\label{cross-domain}
  \centering
  \small
  \begin{tabular}{l*{5}{c}}
    \toprule
    \multirow{2}{*}{Method} & \multicolumn{1}{c}{Source} & \multicolumn{4}{c}{Target} \\
    \cmidrule (lr){2-2} \cmidrule (lr){3-6}
    \addlinespace[1pt]
    & \multicolumn{1}{c}{ImageNet} & \multicolumn{1}{c}{-V2} & \multicolumn{1}{c}{-Sketch} & \multicolumn{1}{c}{-A} & \multicolumn{1}{c}{-R} \\
    \midrule
    CoOp & 71.51 & 64.20 & 47.99 & 49.71 & 75.21 \\
    % Spotlighter & 72.17 &  69.62 & 50.18 & \textbf{69.83} & 83.56 \\
    LwEIB & 71.31 & 64.47 & 50.07 & 51.00 & 77.81 \\
    TAC & \textbf{72.77} & 65.97 & 50.30 & 51.73 & 78.50 \\
    \midrule
    \rowcolor{orange!10}
   CAPT & 72.69 & \textbf{69.71} & \textbf{50.46} & \textbf{57.73} & \textbf{83.61} \\
    \bottomrule
  \end{tabular}
\end{table}
\section{Experiments}
\subsection{Experimental Setup}
\noindent\textbf{Datasets.}
\label{data}
We follow the standard evaluation protocol established in previous studies~\cite{ZhouYLL22,ZhouYL022} to conduct base-to-new and cross-datasets experiments across 11 benchmark datasets, \emph{i.e.,} ImageNet~\cite{DengDSLL009}, Caltech~\cite{Fei-FeiFP07}, OxfordPets~\cite{ParkhiVZJ12}, StanfordCars~\cite{Krause0DF13}, Flowers~\cite{NilsbackZ08}, Food101~\cite{BossardGG14}, FGVCAircraft~\cite{MajiRKBV13}, EuroSAT~\cite{HelberBDB19}, UCF101~\cite{abs-1212-0402}, DTD~\cite{CimpoiMKMV14}, and SUN397~\cite{XiaoHEOT10}. For cross-domain generalization, we experiment on ImageNet-V2~\cite{recht2019imagenet}, ImageNet-Sketch~\cite{wang2019learning}, ImageNet-A~\cite{hendrycks2021nae} and ImageNet-R~\cite{hendrycks2021many}.

\noindent\textbf{Baselines.}
We compare with many state-of-the-art methods, including 
CoOp~\cite{ZhouYLL22},
MaPLe~\cite{MAPLE23},
PromptKD~\cite{li2024promptkd},
Spotlighter~\cite{gao2025spotlighter},
2SFS~\cite{farina2025rethinking},  
TAC~\cite{hao2025task},
LwEIB~\cite{yang2025learning},
TAP~\cite{ding2024tree},
to validate the superiority of CAPT.

\begin{table}[t]
\caption{Comparison with other methods on the few-shot learning setting with average accuracy.}
\label{few-shots}
\centering
\small
\begin{tabular}{c|ccccc}
\hline
\multirow{2}{*}{Method} & \multicolumn{5}{c}{Shot} \\
& 1 & 2 & 4 & 8 & 16 \\
\hline
CoOP & 68.09 & 70.13 & 73.59 & 76.45 & 79.01 \\
MaPLe &61.79 &65.28 &70.66 &73.82 &78.55 \\
PromptKD &72.47&75.19&78.46 &79.56 &86.96\\
 % Spotlighter &  72.53 &  75.76 &  \textbf{78.80} &  81.81 &  85.65 \\
 2SFS  &72.55 & 73.28 &78.50&80.7&85.50\\
      \midrule
     \rowcolor{orange!10}
 CAPT &\textbf{73.11} & \textbf{75.83} & \textbf{78.69} &\textbf{81.91} &\textbf{87.41}\\
\hline
\end{tabular}
\end{table}
\noindent\textbf{Implementation Details.}   
We adopt the ViT-B/16 CLIP model to conduct all of our experiments.
We report accuracies on base and novel classes along with their harmonic mean (HM), averaged over three independent runs. To ensure fairness, all results are averaged across three random seeds to mitigate randomness.
Following standard prompt-tuning settings, we first collect misclassified samples from the base model to construct the Confusion Bank, and then train Diff-Manner Adapter in SAM and MGDE module while keeping the original backbone frozen. 
In each training iteration, the model selects five confusion sample pairs with a batch size of 4. 
Training is conducted for 25 epochs on all datasets except EuroSAT, which is trained for 50 epochs to better capture its higher intra-class variability and complex spatial patterns.
We employ PromptKD~\cite{li2024promptkd} to construct the Confusion Bank, while results based on other models are detailed in the Appendix~\ref{bank}. 
All experiments are conducted on a single NVIDIA RTX 4090 GPU.
% Adhering to the standard configurations of prompt-tuning baselines, we first obtain confusion samples and mistake statistics, followed by fine-tuning the backbone with the proposed XXXX.
% The batch size is set to 32, with 5 confusion samples selected and 3 experts employed in the MoE.
% Meanwhile, the training is conducted for 25 epochs on all datasets, except for EuroSAT, which is trained for 50 epochs to capture its higher complexity better.
% For a fair comparison, we report accuracies on both base and novel classes, and their harmonic mean averaged over 3 independent runs.
% In all experiments, we employ PromptKD to obtain misclassified and confusing category information, while the use of other models for confusion analysis is discussed in detail in the Appendix.
\subsection{Experimental Results}
\noindent\textbf{Base-to-Novel Generalization.}
Table \ref{tab:base-to-new-generalization} presents the base-to-novel generalization results across 11 datasets.
By learning confusion relationships, our method achieves consistent improvements across all evaluation metrics (Base, New, and HM), outperforming existing approaches.
Specifically,  CAPT achieves 87.41\% accuracy on base classes and 80.90\% on novel classes, demonstrating that it can exploit confusion information to refine feature representations from its own prediction misalignment, enhancing sensitivity to fine-grained semantics and overall generalization. 
This boosts performance on novel classes and strengthens the connections between known and unknown categories, facilitating understanding and adaptation to unseen classes.
 
\noindent\textbf{Cross-Dataset Transfer and Domain  Generalization.}
We train ImageNet on all classes as the source and evaluate the model on the datasets mentioned in \ref{data} under a zero-shot setting for both cross-dataset and cross-domain tasks.
Table~\ref{tab:cross-dataset} and Table~\ref{cross-domain} report the corresponding results, demonstrating that the learned confusion information can be effectively transferred across datasets and domains, exhibiting strong generalizability and consistently preserving the model’s performance on unseen data distributions.

 \noindent\textbf{Few-shot Classification.}
Adhering to the baselines, we used 1/2/4/8/16-shot settings for training and calculated the accuracy on 11 datasets.
Table~\ref{few-shots} shows that when compared with other methods,  CAPT displays overall consistent improvement among all settings, demonstrating robustness and efficacy even in challenging low-data regimes.

\begin{figure} 
\includegraphics[width=1\linewidth]{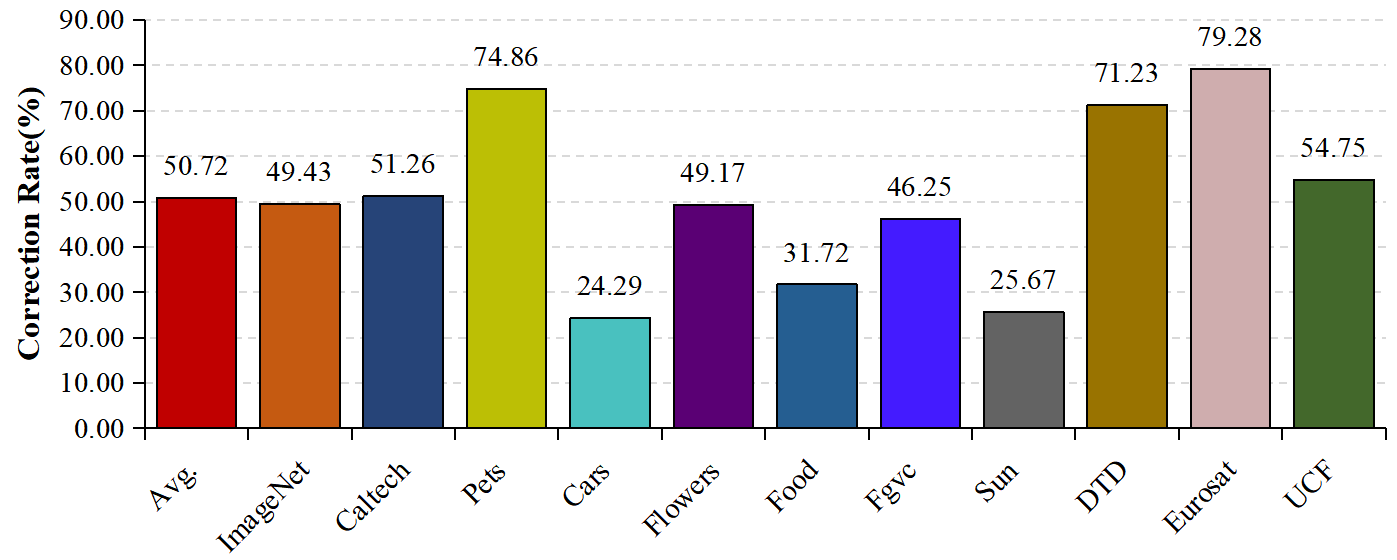} 
\caption{Correction Rate of misclassified samples stored in Confusion Bank.}
% \vspace{-1.7em}  
% \caption {\textbf{Motivation}. (a) Instances of model misclassification, where errors occur in fixed categories. (b) Proportion of misclassified samples within each category, as not all samples are equally prone to confusion. (c) Grad-CAM visualizations of confused samples, model either lacks global perception or overlooks details. }  
\label{corret}  
\vspace{-1em}
\end{figure}

 \noindent\textbf{Correction Rate.}
 Figure~\ref{corret} illustrates the correction of misclassified samples stored in the Confusion Bank. An average correction rate of 50.72\% demonstrates that CAPT can effectively learn confusing knowledge from a small set of selected confusing samples and enable finer-grained recognition of the most easily confusable categories.

\begin{table}[t]
\centering
\caption{Ablation study of components in  CAPT on base-to-new tasks.~~~SEM: Semantic Confusion Miner.~~~SAM: Sample Confusion Miner.~~~MGDE: Multi-Granularity Discrepancy Expert.  }
\begin{tabular}{c c c | c c c}
\toprule
SEM &  SAM &  MGDE 
& Base &  Novel & HM \\
\midrule
\checkmark &    &   & 80.69 & 64.39 & 71.62 \\
 & \checkmark &  & 83.28 & 63.41 & 72.00 \\
\checkmark   & \checkmark & & 85.23 & 65.17 &73.86   \\
\rowcolor{orange!10}
\checkmark & \checkmark & \checkmark  &\textbf{87.41} &\textbf{80.90} &\textbf{83.90} \\
\bottomrule
\end{tabular}
\label{tab:ablation}
\end{table}

 \subsection{Ablation Study}
\noindent\textbf{Validity of Proposed Components.}    
\begin{table}[h]
\setlength{\tabcolsep}{4pt} 
    \centering
    \small
    \caption{Effects of the Confusion Statistics in Confusion Score. }
    \begin{tabular}{c|ccc}
        \toprule
        Method & Base & Novel & HM \\
        \hline
         CAPT~$w/o$  Confusion Statistics.& 86.77 & 78.84  &   82.62 \\
       \rowcolor{orange!10}
           CAPT $w$ Confusion Statistics. & \textbf{87.41} &  \textbf{80.90}&  \textbf{83.90}  \\
        \bottomrule
    \end{tabular}
    \label{tab:semantic}
\end{table}
\begin{table}[h]
\setlength{\tabcolsep}{4pt} 
    \centering
    \small
    \caption{Confusion Pair Selection with  Pseudo-GT and Real-GT. }
    \begin{tabular}{c|ccc}
        \toprule
        Method & Base & Novel & HM \\
        \hline
       Real-GT & 86.94 & 79.52 &  83.06  \\
      \rowcolor{orange!10}
         Pseudo-GT&  \textbf{87.41}&  \textbf{80.90} &  \textbf{83.90}   \\
        \bottomrule
    \end{tabular}
    \label{tab:initialization}
\end{table}
Table~\ref{tab:ablation} highlights the contributions of the proposed SEM, SAM, and MGDE modules.
Mining confusion information solely at the semantic level significantly reduces accuracy on base classes, while mining solely at the sample level severely impairs generalization on novel classes. Only by effectively integrating confusion information from both semantic and sample levels (MGDE) can the model fully capture the complex confusion relationships between categories, achieving more accurate classification and stronger generalization.

\noindent\textbf{Effects of  Confusion Statistics in 
Calculating Confusion Score.}
In the Semantic Confusion Miner, instead of directly selecting samples with the highest confidence, we 
compute a more discriminative Confusion Score by integrating the confidence of the current sample with global Confusion Statistics.
As shown in Table~\ref{tab:semantic}, this strategy achieves notably higher accuracy compared to the variant without Confusion Statistics. This demonstrates that leveraging the distributional characteristics of the global semantic space not only mitigates the risk of confusion information being overly concentrated but also enables the acquisition of more representative confusion relationships, allowing the model to comprehensively capture inter-class confusion patterns and refine semantic boundaries.

\noindent\textbf{Confusion Pair Selection with  Pseudo-GT and Real-GT.}
 \begin{figure} 
\includegraphics[width=1\linewidth]{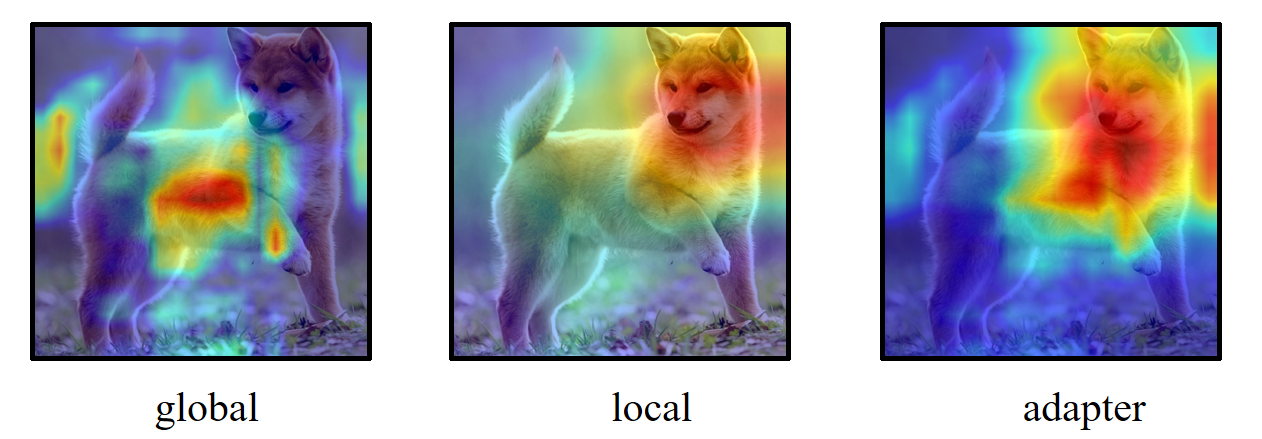} 
\caption{Grad-Cam of the effects of Diff-Manner Adapter, only using the global part, the local part, and the whole.}
\label{fig:grad_ca}  
\vspace{-1em}
\end{figure}
In the Semantic Confusion Miner, we treat the class with the highest confidence as the pseudo ground truth for the current sample to simulate potential confusion behaviors in the semantic discrimination process, rather than directly using the true label. 
As shown in Table~\ref{tab:initialization}, the comparison results demonstrate that explicitly modeling such confusion behaviors enables the model to better capture the inherent uncertainty in semantic decision-making, leading to more discriminative and robust confusion feature representations.
% \begin{table}[h]
% \setlength{\tabcolsep}{4pt} 
%     \centering
%     \small
%     \caption{Effects of the Confusion Instance Sampling.}
%     \begin{tabular}{c|cc}
%         \toprule
%         Method   & FPS & HM \\
%         \hline
%         PromptKD  & 79.55 &  83.04  \\
%          DPC &  \textbf{80.90} &  \textbf{83.90}   \\
%         \bottomrule
%     \end{tabular}
%     \label{tab:effen}
% \end{table}
\noindent\textbf{Effects of Diff-Manner Adapter in Capturing Confusion Cues.}
Figure~\ref{fig:grad_ca} presents the Grad-CAM~\cite{selvaraju2017grad} visualizations obtained from the Diff-Manner Adapter, where only global, only local, and both types of features are retained.
It can be observed that focusing solely on either global or local features tends to overlook sample-level confusion cues, hindering the discrimination of visually or semantically similar categories.
In contrast, leveraging both enables the model to capture holistic semantics and key confusion patterns, achieving finer differentiation of confusable samples.

\noindent\textbf{Analysis of Inference Efficiency.}
During inference, no confusing samples need to be indexed, and the Diff-Manner Adapter in SAM together with MGDE can already effectively encode confusion information.
Compared with the baseline (PromptKD), our method introduces only a \textbf{323.56} FPS overhead, achieving \textbf{2591.73 }FPS, and is still \textbf{343.56} FPS faster than DPC built on the same baseline.

\noindent\textbf{Effect of Choice Confusion Instance Sampling in SAM.}
Confusing samples are retrieved via Semantic Confusion Pairs from the Confusion Bank instead of being randomly selected from the original dataset. 
As shown in Table~\ref{tab:representative}, this strategy yields superior results, as semantically indexed samples better reflect intrinsic error patterns and inter-class ambiguities, while random sampling introduces irrelevant noise that weakens confusion discrimination.

\begin{table}[h]
\setlength{\tabcolsep}{4pt} 
    \centering
    \small
    \caption{Effects of the Confusion Instance Sampling.}
    \begin{tabular}{c|ccc}
        \toprule
        Method & Base & Novel & HM \\
        \hline
        Random Choice & 86.86 & 79.55 &  83.04  \\
      \rowcolor{orange!10}
        Representative Samples&  \textbf{87.41}&  \textbf{80.90} &  \textbf{83.90}   \\
        \bottomrule
    \end{tabular}
    \label{tab:representative}
\end{table}

% \noindent\textbf{Effects of different Confusion Pair and Representative Sample Number.}
% Figure~\ref{number} illustrates the influence of the number of representative samples in the Sample Confusion Miner and confusion pairs in the Semantic Confusion Miner.
% We observe that retaining one representative sample per confusing category yields the best performance, as excessive samples introduce redundant confusion information, making it harder for the model to learn effectively. For confusion pairs, the best results are achieved when five pairs are retained; too few pairs fail to capture sufficient confusion information, while too many introduce unnecessary noise that hinders the model's confusion learning.

%  \begin{figure} 
% \includegraphics[width=1\linewidth]{./number.png} 
% \caption{Effects of Different Confusion Pair in SEM and Representative Sample Number in SAM.}
% % \vspace{-1.7em}  
% % \caption {\textbf{Motivation}. (a) Instances of model misclassification, where errors occur in fixed categories. (b) Proportion of misclassified samples within each category, as not all samples are equally prone to confusion. (c) Grad-CAM visualizations of confused samples, model either lacks global perception or overlooks details. }  
% \label{number}  
% \vspace{-1em}
% \end{figure}
 
\section{Conclusion}
\vspace{-0.5em}
We introduce CAPT, a confusion-aware prompt tuning framework that enables vision–language models to learn from their own misalignment by explicitly modeling confusion relationships at both semantic and sample levels. We observe that certain categories are consistently misclassified into specific others, forming fixed confusion patterns. To address this, CAPT constructs a Confusion Bank and employs Semantic and Sample Confusion Miners, together with a Multi-Granularity Discrepancy Expert module, to capture confusion cues across multiple granularities and generate more discriminative, robust representations. Experiments on 11 benchmarks demonstrate consistent improvements on both base and novel classes, resolving 50.72\% of confusable samples. 
Our work positions confusion modeling as a promising direction for self-corrective, fine-grained vision–language learning, with future work on task-adaptive confusion mining for dynamic ambiguity.
% We introduce xxxx, a confusion-aware prompt tuning framework that enables vision-language models to learn from their own mistakes by explicitly modeling confusion relationships at both semantic and sample levels. We find that some categories are consistently mispredicted as specific other categories, forming a “fixed confusion pattern.” Our approach leverages the Semantic Confusion Miner, Sample Confusion Miner, and MGDE module to effectively capture multi-granularity confusion cues, producing more discriminative and robust representations. Extensive experiments demonstrate consistent gains in both base and novel classes, resolving 50.72\% of confusable samples. This work highlights confusion modeling as a promising direction for self-corrective and fine-grained vision-language learning, paving the way for future exploration in open-world and multi-modal reasoning.

\section*{Acknowledgments}

This work was supported by the Major Projects of the National Natural Science Foundation of China (Grant Nos.~72293580 and 72293583); 
the Hainan Provincial Joint Project of Lian International Education Innovation Pilot Zone (Grant No.~624LALH002); 
and the Foundation of Key Laboratory of Big Data \& Artificial Intelligence in Transportation (Beijing Jiaotong University), 
Ministry of Education (No.~BATLAB202403).

{
\small
\bibliographystyle{ieee_fullname}
\bibliography{egbib}
}

\newpage
\appendix

\twocolumn[
\begin{center}
    {\Large \textbf{Supplementary Material of \\ 
    ~~~~~~~~~\\
    \textit{CAPT: Confusion-Aware Prompt Tuning for Reducing Vision-Language
    Misalignment}}}
\end{center}
]
\section{Discussion}
 CAPT explicitly models systematic confusion at both the semantic and sample levels, enabling the model to extract stable confusion structures from its own misclassifications. However, on relatively easier datasets such as Flowers~\cite{NilsbackZ08} and Food101~\cite{BossardGG14}, where overall accuracy is \textbf{already high} and genuinely confusable samples are \textbf{scarce}, the room for improvement is limited, resulting in modest performance gains.
 In contrast, CAPT yields more pronounced gains on more complex datasets such as UCF101~\cite{abs-1212-0402}, DTD~\cite{CimpoiMKMV14}, with about \textbf{2.35\%} improvement.
  Nevertheless, CAPT achieves stable superiority across all 11 benchmarks, with a 50.72\% correction rate on confusing samples, demonstrating its effectiveness on real challenging classes.
%   By leveraging global confusion statistics and pseudo-labels, CAPT mitigates noise accumulation, ensuring robust and generalizable self-supervised confusion signals.
% This demonstrates that CAPT not only improves accuracy on challenging categories but also provides a general framework for leveraging model-inherent confusions in a stable and effective manner.
\section{Impact of Model Choice on Confusion Bank Construction}
\label{bank}
In the main text, we use PromptKD to construct our confusion bank. In practice, different models often exhibit similar and stable confusion patterns across many categories. However, due to differences in their feature spaces, alignment behaviors, and error distributions, confusion banks built from different models vary significantly in the coverage and representativeness of confusing instances, which in turn leads to different downstream effects.
delivers the highest correction rate.
Shown in Table~\ref{tab:bank}, PromptKD achieves the highest overall accuracy, whereas the Confusion Bank built from CLIP
delivers the highest correction rate.
Thus, stronger models help construct more accurate recognition benchmarks, while simpler models tend to capture more transferable and consistent confusion patterns that are more effective for error correction.
\begin{table}[h]
\setlength{\tabcolsep}{4pt} 
    \centering
    \small
    \caption{ Impact of Model Choice on Confusion Bank Construction. }
    \begin{tabular}{c|ccc|c}
        \toprule
        Method Choice & Base & Novel & HM & Correction Rate \\
        \hline
       CLIP & 73.56 & 77.23 & 77.35  &64.28 \\
      
        CoOp &  84.19& 71.88 & 77.55 &42.71   \\

        MaPLe & 85.64& 79.14 &82.26 &39.39 \\
        TAC &84.41 &78.67 &81.44 & 49.20\\
        PromptKD &87.41 &80.90 &83.90 &50.72\\
        \bottomrule
    \end{tabular}
    \label{tab:bank}
\end{table}
\begin{figure}
    \centering
    \includegraphics[width=1.0\linewidth]{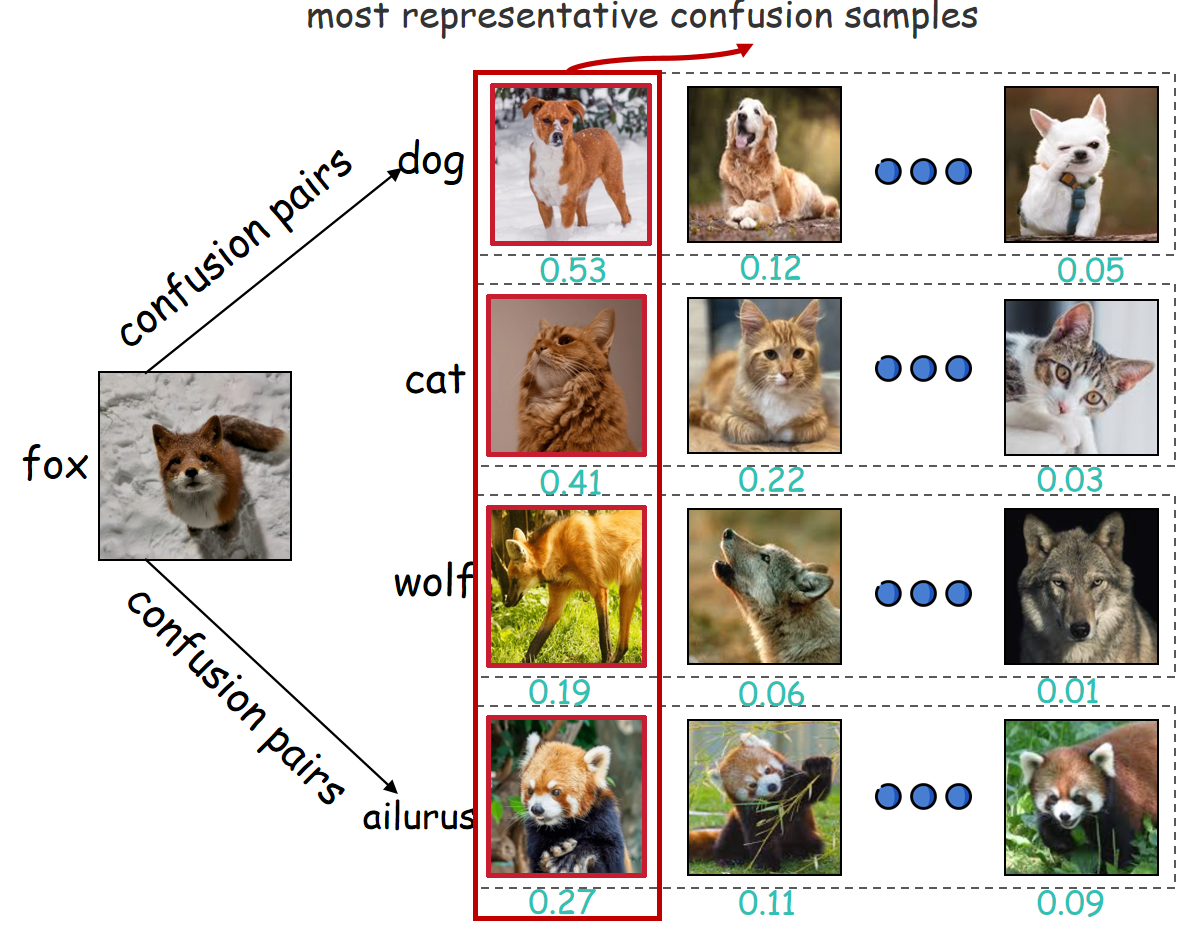}
    \caption{Instance of selecting representative confusion samples.}
    \label{fig:represent}
\end{figure}
\section{More Details about the Most Representative Confusion Samples}
In the Sample Confusion Miner (SAM) module, we leverage the previously identified semantic confusion pairs and apply Eq~\ref{sim} to retrieve, for each confusable class, the most representative confusing sample.
Figure~\ref{fig:represent} illustrates examples of selecting representative confusion samples for specific instances. For each confusable class, only the most similar sample is retained as the representative confusing example, which helps filter out noisy candidates, sharpen the model’s focus on critical confusion patterns, and ultimately enhance its ability to discriminate fine-grained differences and make more stable predictions.

\section{Datasets}
To obtain a comprehensive understanding of our method’s capability, we evaluate it across an extensive suite of benchmarks covering standard recognition, transfer learning, and distribution shift scenarios, as summarized in Table~\ref{tab:dataset}. Our core evaluation is conducted on 11 widely adopted datasets that span a broad range of visual domains—including general object recognition (ImageNet~\cite{DengDSLL009}, Caltech~\cite{Fei-FeiFP07}), fine-grained categorization (OxfordPets~\cite{ParkhiVZJ12}, StanfordCars~\cite{Krause0DF13}, Flowers~\cite{NilsbackZ08}, Food101~\cite{BossardGG14}, FGVCAircraft~\cite{MajiRKBV13}), satellite-based scene understanding (EuroSAT~\cite{HelberBDB19}), large-scale scene recognition (SUN397~\cite{XiaoHEOT10}), human action recognition (UCF101~\cite{abs-1212-0402}), and texture classification (DTD~\cite{CimpoiMKMV14}). These datasets collectively cover a wide spectrum of visually confusable scenarios, ranging from highly similar fine-grained categories to semantically overlapping scene types, making them well-suited for evaluating confusion-aware modeling.
To further assess the adaptability of our approach, we also incorporate three widely used transfer-learning datasets, which provide additional settings where category-level similarity and domain-specific ambiguity introduce new confusion patterns that challenge representation robustness. Beyond standard benchmarks, we additionally examine robustness under real-world distribution shifts using four challenging ImageNet variants: ImageNet-V2~\cite{recht2019imagenet}, ImageNet-Sketch~\cite{wang2019learning}, ImageNet-A~\cite{hendrycks2021nae}, and ImageNet-R~\cite{hendrycks2021many}. These variants introduce complementary perturbations—such as re-sampled images, sketch abstractions, natural adversarial examples, and artistic renditions—that often amplify inherent confusability among visually similar categories.
Collectively, this diverse suite of benchmarks offers a rigorous and multi-dimensional testbed for evaluating not only the overall effectiveness and generalization of our method but also its ability to remain robust under semantically and visually confusing conditions.

\section{Details of Semantic Prompt Generation in Semantic Confusion Miner.}
 \begin{figure} 
\includegraphics[width=1\linewidth]{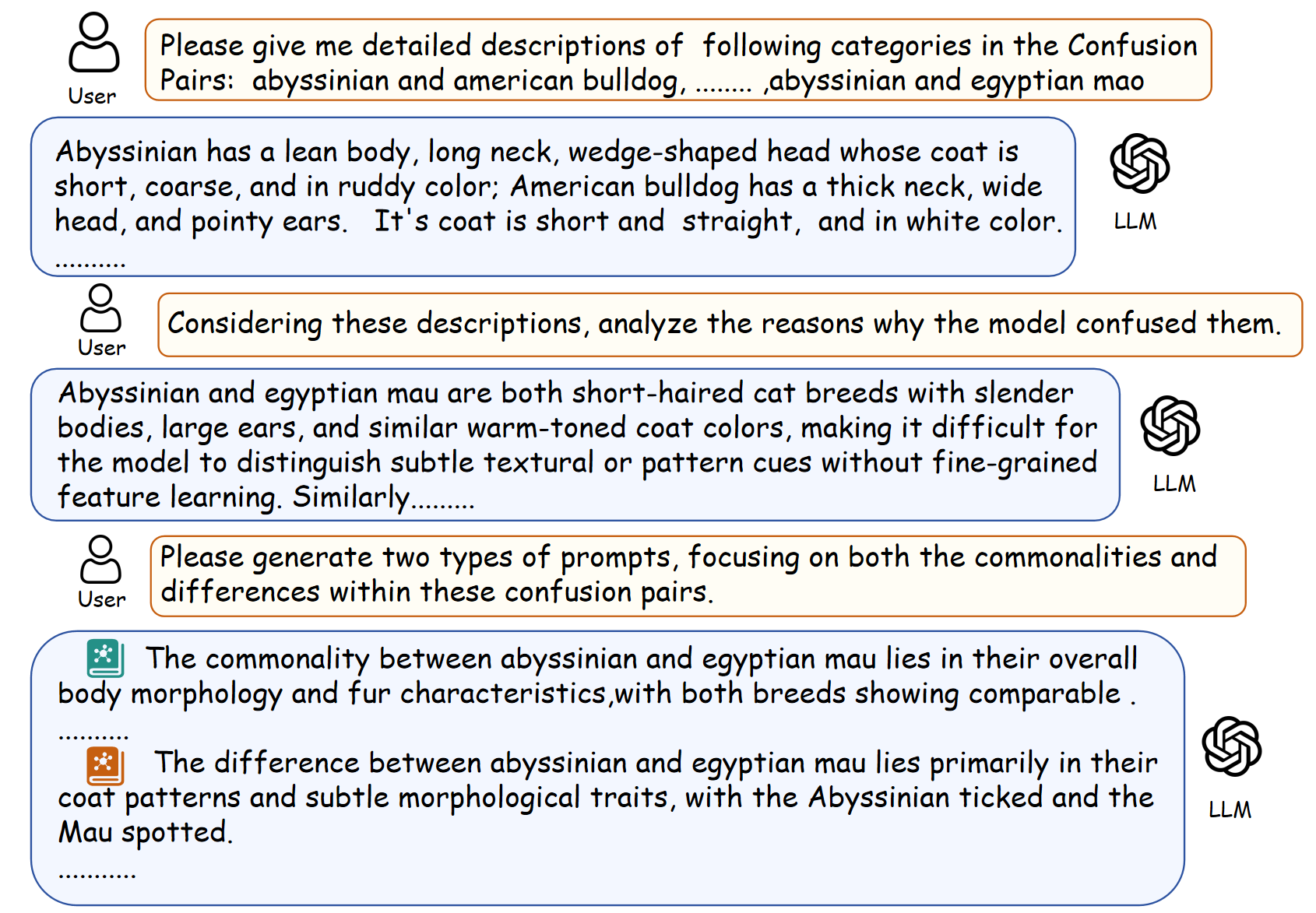} 
\caption{Procedure of generating semantic commonality and difference prompts.}
\label{fig:cot}  
\vspace{-1em}
\end{figure}
We leverage confusion pairs generated by SEM to construct semantic level commonality and difference prompts, which are then used to initialize the expert layers in MGDE, effectively integrating multi-class confusion information. Specifically, as illustrated in the Figure~\ref{fig:cot}, we employ a CoT-based~\cite{wei2022chain,li2024atprompt} procedure to guide prompt generation: the model first produces detailed feature descriptions for the categories involved in the confusion pairs, then uses these descriptions to reason about the 
underlying causes of misalignment, and finally generates the corresponding difference and commonality prompts. By progressively guiding the model through this process, we can more accurately capture the confusion characteristics within the pairs, thereby enhancing the model’s understanding and discrimination of inter-class confusion relationships.

\section{Details of Dynamic Weight in Diff-Manner Adapter }
\begin{figure*} 
\includegraphics[width=1\linewidth]{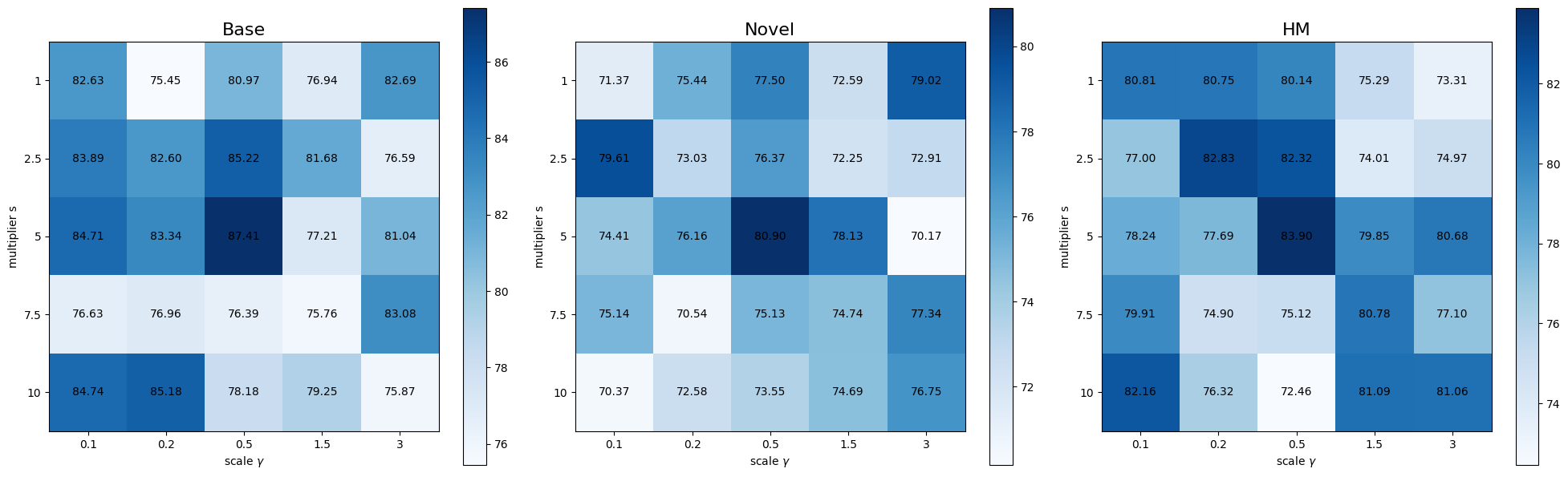} 
% \caption{Heatmaps of model misclassifications across more datasets prove the fixed confusion pattern.}
\caption{ Effects of different $s$ and $\gamma$ in choosing adaptive $\alpha$.}
\label{fig:alpha}  
\vspace{-1em}
\end{figure*}

In Equal~\ref{eq3}, the hyper-parameter $\alpha$ plays a critical role in learning sample confusion features.
On the one hand, when $\alpha$ is too small, the model focuses only on coarse, global confusion structures and fails to align finer-grained confusion cues. On the other hand, an excessively large $\alpha$ leads the model to overemphasize local details, reducing generalization ability and causing severe overfitting.
To balance global confusion patterns and instance-level cues, we introduce an adaptive $\alpha$ scheduling mechanism that enables the model to automatically determine its level of attention based on the confusion intensity of each sample.
We first get the confusion intensity $c_i$ of each representative confusion sample based on Equal~\ref{sim}.
Then,  we compute a dynamic $\alpha$:
\begin{equation}
    \alpha_i =  s \cdot c_i^{\gamma},
\end{equation}
where $s$ and $\gamma$ are new introduced hyper-parameters,  with $s>1$  and $\gamma>0$ to
 control the sensitivity of $\alpha$ to confusion strength.
 In this work, we set $s=5$ and $\gamma=0.5$ and the results of different $s$ and $\gamma$ is shown in the Figure~\ref{fig:alpha}.

 \section{Additional Materials on the Diff-Manner Adapter in SAM}
\begin{figure} 
\includegraphics[width=0.95\linewidth]{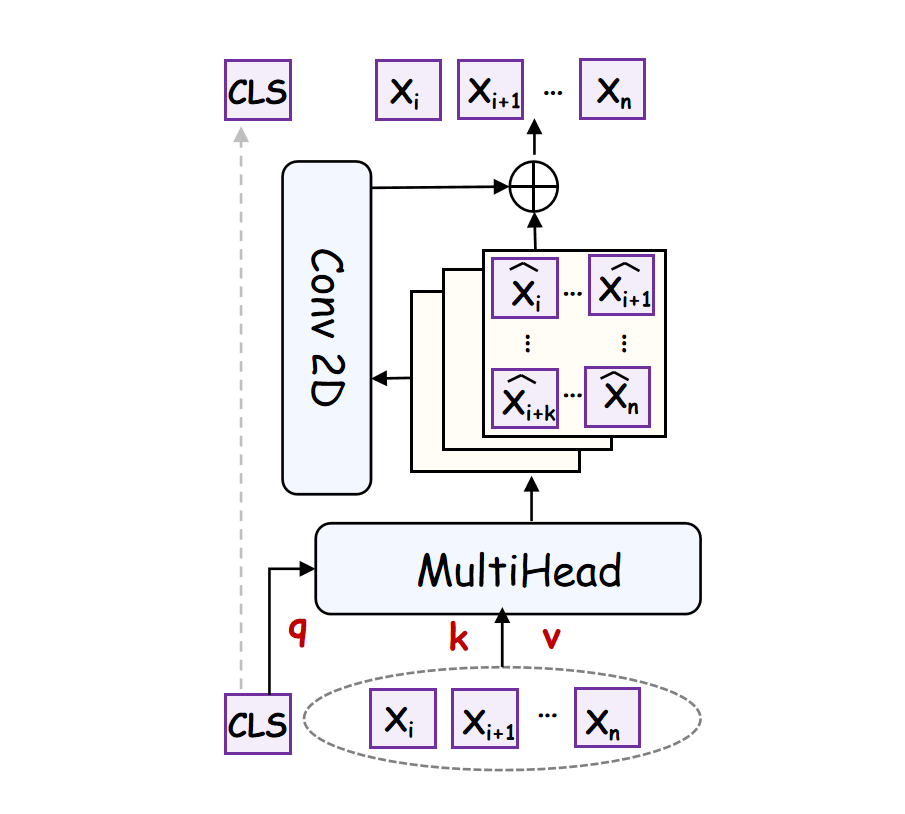} 
\centering
% \caption{Heatmaps of model misclassifications across more datasets prove the fixed confusion pattern.}
\caption{ Framework of Equal~\ref{eq_x},Equal~\ref{eq3} in Diff-Manner Adapter.}
\label{fig:conv}  
\vspace{-1em}
\end{figure}

 We introduce the Diff-Manner Adapter in the Sample Confusion Manner (SAM), which adaptively captures typical confusion features from representative confusion samples from both global and local perspectives. As shown in Figure~\ref{fig:grad_ca}, Grad-CAM visualizations are provided for global-only, local-only, and fused feature representations, with corresponding accuracy results summarized in the accompanying table~\ref{tab:glob}. It is noteworthy that when only global features are used, the accuracy of base classes drops significantly, while using only local features leads to a clear decline in novel class accuracy. This indicates that global and local features play complementary roles in supporting the classification decisions for base and novel classes.
Framework of  Equal~\ref{eq_x},Equal~\ref{eq3} when capturing local sample confusion feature is shown in Figure ~\ref{fig:conv}.
\begin{table}[h]
\setlength{\tabcolsep}{4pt} 
    \centering
    \small
    \caption{Impact of global and local part in Diff-Manner Adapter.  }
    \begin{tabular}{c|ccc}
        \toprule
        Method & Base & Novel & HM  \\
        \hline
      global  & 81.22 & 76.23 &  78.65   \\
      
        local &  85.19& 72.88 &  78.56   \\

     global+local & 87.41& 80.90 &83.90   \\
        \bottomrule
    \end{tabular}
    \label{tab:glob}
\end{table}

\section{Effects of Different Confusion Pair and Representative Sample Number.}
\begin{figure} 
\includegraphics[width=1\linewidth]{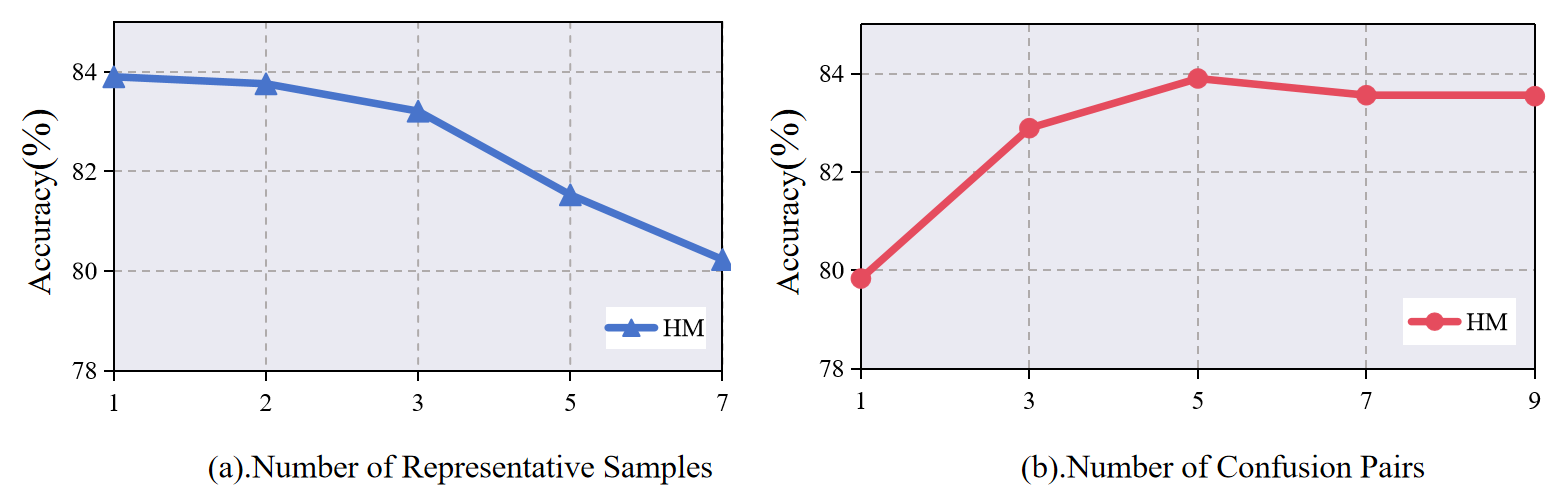} 
\caption{Effects of Different Confusion Pair in SEM and Representative Sample Number in SAM.}
% \vspace{-1.7em}  
% \caption {\textbf{Motivation}. (a) Instances of model misclassification, where errors occur in fixed categories. (b) Proportion of misclassified samples within each category, as not all samples are equally prone to confusion. (c) Grad-CAM visualizations of confused samples, model either lacks global perception or overlooks details. }  
\label{number}  
\vspace{-1em}
\end{figure}

Figure~\ref{number} illustrates the influence of the number of representative samples in the Sample Confusion Miner and confusion pairs in the Semantic Confusion Miner.
We observe that retaining one representative sample per confusing category yields the best performance, as excessive samples introduce redundant confusion information, making it harder for the model to learn effectively. For confusion pairs, the best results are achieved when five pairs are retained; too few pairs fail to capture sufficient confusion information, while too many introduce unnecessary noise that hinders the model's confusion learning.

\section{Further Evidence of the Fixed Confusion Pattern}
\begin{figure*} 
\includegraphics[width=1\linewidth]{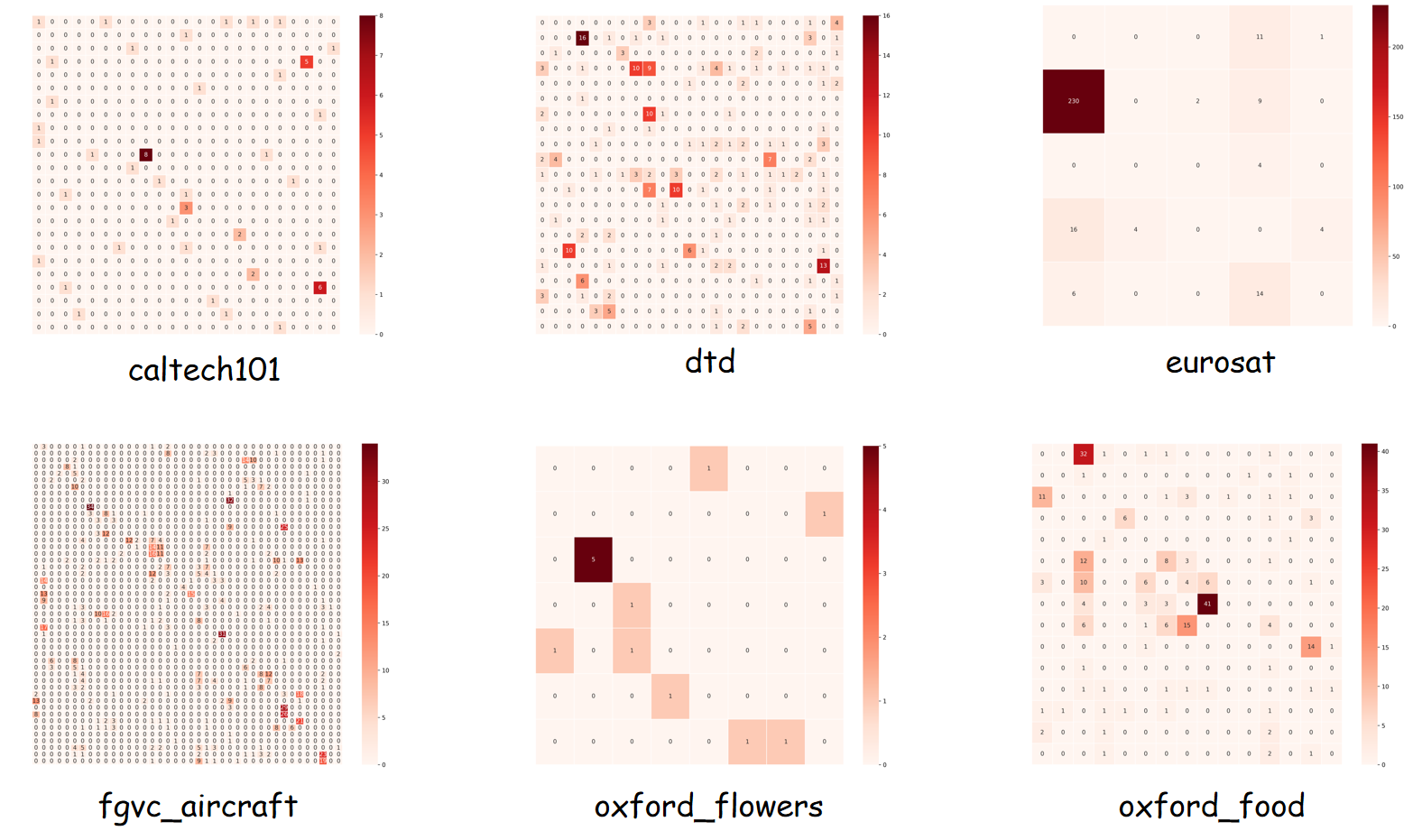} 
\caption{Heatmaps of model misclassifications across more datasets prove the fixed confusion pattern.}
\label{fig:conf}  
\vspace{-1em}
\end{figure*}

We observe a fixed confusion pattern in which certain categories are consistently misaligned into specific target classes with significantly higher probability. This observation provides a foundation for CAPT to address the alignment problem.
Figure~\ref{fig:com} illustrates the confusion pattern observed on the Oxford dataset. This finding is further substantiated by the misalignment analysis across multiple datasets in Figure~\ref{fig:conf}.
For instance, in the dataset Caltech101, 'cougar\_body' is misclassified as 'cougar\_face' 8 times with almost no other errors; in the dataset Food101, 'cake' is misclassified as 'chocolate' 41 times exclusively. This pattern is further underscored by a striking example from the dataset FGVCAircraft, where 'Permanent Crop Land' is misclassified as 'Pasture Land' 230 times—a frequency orders of magnitude higher than other single-digit error rates. These results collectively demonstrate that the fixed confusion pattern is not an isolated occurrence, but a pervasive systemic bias, revealing a fundamental limitation in current models for fine-grained recognition.

\section{Impact of Confusing Sample Quality}
To further evaluate the model’s sensitivity to misclassified samples, we randomly injected 5\% and 10\% Gaussian noise into the selected most representative confusion samples.
Results are reported in the Table~\ref{tab123}.
We observe that introducing a small amount of noise leads to only marginal performance degradation. This not only demonstrates the robustness of our method to perturbations, but also indicates that our statistics-based design effectively mitigates the impact of sample noise.

 \begin{table}[ht]
\setlength{\tabcolsep}{2pt} 
    \centering
    \small
    \caption{Effects of Noise Add.}
     \vspace{-1.0em}
    \begin{tabular}{c|ccc}
        \toprule
        Method & Base & Novel & HM \\
        \hline
        
                 Random Add 5\% Gaussian Noise  &79.87 &74.49 &77.09\\
                 Random Add 10\% Gaussian Noise &72.41&69.56   &70.96\\
 
        \bottomrule
    \end{tabular}
    \label{tab123}
\end{table}

 \begin{table*}[h]
\centering
\setlength{\tabcolsep}{3pt}
\caption{Statistics of the datasets used in our experiments}
\label{tab:dataset}
\resizebox{\textwidth}{!}{
\begin{tabular}{l|c|c|c|c|c|c}
\toprule
\textbf{Dataset} & \textbf{Classes} & \textbf{Train} & \textbf{Val} & \textbf{Test} & \textbf{Task} & \textbf{Hand-crafted Prompt} \\
\midrule
ImageNet        & 1,000 & 1.28M & N/A  & 50,000 & General object recognition & ``a photo of a [CLASS].''\\ 
Caltech101      & 100   & 4,128 & 1,649 & 2,465 & General object recognition & ``a photo of a [CLASS].'' \\
EuroSAT         & 10    & 13,500 & 5,400 & 8,100 & Remote sensing classification & ``a satellite image of [CLASS].''  \\
SUN397          & 397   & 15,880 & 3,970 & 19,850 & Scene classification & ``a photo of a [CLASS].'' \\
DTD             & 47    & 2,820 & 1,128 & 1,692 & Texture classification & ``a [CLASS] texture.'' \\
UCF101          & 101   & 7,639 & 1,808 & 3,783 & Human action recognition & ``a photo of a person doing [CLASS].'' \\
FGVCAircraft    & 100   & 3,334 & 3,333 & 3,333 & Fine-grained aircraft recognition & ``a photo of a [CLASS], an aircraft type.'' \\
OxfordPets      & 37    & 2,944 & 736  & 3,669 & Fine-grained pet recognition & ``a photo of a [CLASS], a pet breed.'' \\
StanfordCars    & 196   & 6,509 & 1,635 & 8,041 & Fine-grained car recognition & ``a photo of a [CLASS], a car model.'' \\
Flowers102      & 102   & 4,093 & 1,633 & 2,463 & Fine-grained flower recognition & ``a photo of a [CLASS].''\\
Food101         & 101   & 50,500 & 20,200 & 30,300 & Food image classification & ``a photo of a [CLASS], a type of food.'' \\
\midrule
ImageNet-V2     & 1,000 & N/A & N/A & 10,000 & ImageNet distribution shift & ``a photo of a [CLASS].''\\
ImageNet-Sketch & 1,000 & N/A & N/A & 50,899 & ImageNet distribution shift & ``a sketch of a [CLASS].''\\
ImageNet-A      & 1,000 & N/A & N/A & 7,500 & Natural adversarial examples & ``a photo of a [CLASS].''\\
ImageNet-R      & 1,000 & N/A & N/A & 30,000 & Artistic rendition recognition & ``a rendition of a [CLASS].''\\
\bottomrule
\end{tabular}
}
\end{table*}


\end{document}